# Incivility and Rigidity: Evaluating the Risks of Fine-Tuning LLMs for Political Argumentation

**Svetlana Churina** and **Kokil Jaidka,**
Centre for Trusted Internet & Community,
National University of Singapore,
Singapore

## Abstract

Incivility on platforms such as Twitter (now X) and Reddit complicates the development of AI systems that can support productive, rhetorically sound political argumentation. We present experiments with *GPT-3.5 Turbo* fine-tuned on two contrasting datasets of political discourse: high-incivility Twitter replies to U.S. Congress and low-incivility posts from Reddit's *r/ChangeMyView*. Our evaluation examines how data composition and prompting strategies affect the rhetorical framing and deliberative quality of model-generated arguments. Results show that Reddit-finetuned models generate safer but rhetorically rigid arguments, while cross-platform fine-tuning amplifies adversarial tone and toxicity. Prompt-based steering reduces overt toxicity (e.g., personal attacks) but cannot fully offset the influence of noisy training data. We introduce a rhetorical evaluation rubric - covering justification, reciprocity, alignment, and authority - and provide implementation guidelines for authoring, moderation, and deliberation-support systems.

**Disclaimer** —- This paper contains some profanity that may be disturbing to some readers.

## 1 Introduction

Large language models (LLMs) are increasingly applied in computational social science to emulate, analyze, and generate human discourse. Their capacity for producing coherent, contextually grounded text makes them useful for studying argumentative practices in online political communication. However, using LLMs to support or simulate public deliberation raises concerns about rhetorical fidelity, bias amplification, and discourse degradation - particularly when models are fine-tuned on high-incivility data typical of social media.

Two related risks are salient. First, fine-tuning on highly adversarial data, such as replies to politicians on Twitter (now X), may encode toxic or polarizing rhetorical patterns, reducing the model's ability to generalize to balanced argumentation. Second, recursive training on synthetic model outputs can compound stylistic distortions over time (Shumailov et al., 2024), degrading deliberative quality and reliability.

This study investigates how platform-specific data shape the rhetorical and deliberative properties of AI-generated political arguments. We focus on whether fine-tuned models preserve the reason-giving and reciprocity that characterize productive political dialogue. Our research pursues two objectives:

- **RO1:** Characterize the rhetorical and deliberative quality of political arguments produced by LLMs fine-tuned on high-incivility social media data.
- **RO2:** Evaluate mitigation strategies - such as balanced training and prompt-based steering - for improving deliberative quality in generated arguments.

## 2 Study Contributions

Political arguments are structured, stance-taking texts designed to persuade or contest viewpoints (Bender et al., 2011; Rowe, 2015). Yet not all arguments advance deliberation. We define *deliberative quality* as the degree to which an argument provides reasons, acknowledges opposition, and seeks common ground - features essential to inclusive, respectful discourse.

While prior work has explored argument generation and stance framing, little is known about how platform-specific norms influence the rhetorical behavior of fine-tuned models. Our study bridges this gap through three contributions:

- **Empirical:** We show that fine-tuned models inherit platform-specific rhetorical biases: those trained on high-incivility data (Twitter)[1] produce more adversarial rhetoric, while those trained on low-incivility data (Reddit CMV) yield more civil but stylistically constrained arguments.
- **Methodological:** We introduce an LLM-assisted annotation pipeline for identifying *alignment* and *authority* moves - two key indicators of deliberative engagement.
- **Practical:** We offer design guidelines for configuring models, curating datasets, and crafting prompts that balance rhetorical diversity, civility, and deliberative quality.

Together, these contributions advance the study of rhetoric-aware fine-tuning and provide empirical foundations for designing AI systems that enhance, rather than erode, deliberative discourse.

## 3 Related Work

Research on AI-driven argumentation has examined prompting, factuality, and discourse quality. Figueras and Agerri (2024) and Lin et al. (2023) showed that LLMs can generate concise, deliberative arguments, though persuasiveness varies by framing and ideology (Simmons, 2023; El Baff et al., 2024). However, fine-tuning on low-quality political discourse risks amplifying bias and misinformation (Giarelis et al., 2024; Dykes et al., 2024), with recursive training further degrading diversity and factual integrity (Shumailov et al., 2024).

Efforts to quantify deliberative quality (Behrendt et al., 2024) highlight indices such as civility, rationality, and reciprocity. Building on this work, we shift from classification to generation: using the annotated *CLAPTON* dataset (Jaidka, 2022b), we examine how LLMs internalize and reproduce deliberative features - specifically *justification* and *reciprocity* (Steenbergen et al., 2003). Our analysis thus connects rhetorical modeling with practical design considerations for AI-mediated deliberation.

## 4 Method

We investigate how variations in training data influence the linguistic and stylistic properties of fine-tuned language models. To this end, we conducted comparative analyses of model outputs produced after fine-tuning on datasets curated from different online platforms. This approach allowed us to identify patterns associated with incivility and rhetorical variation across sources.

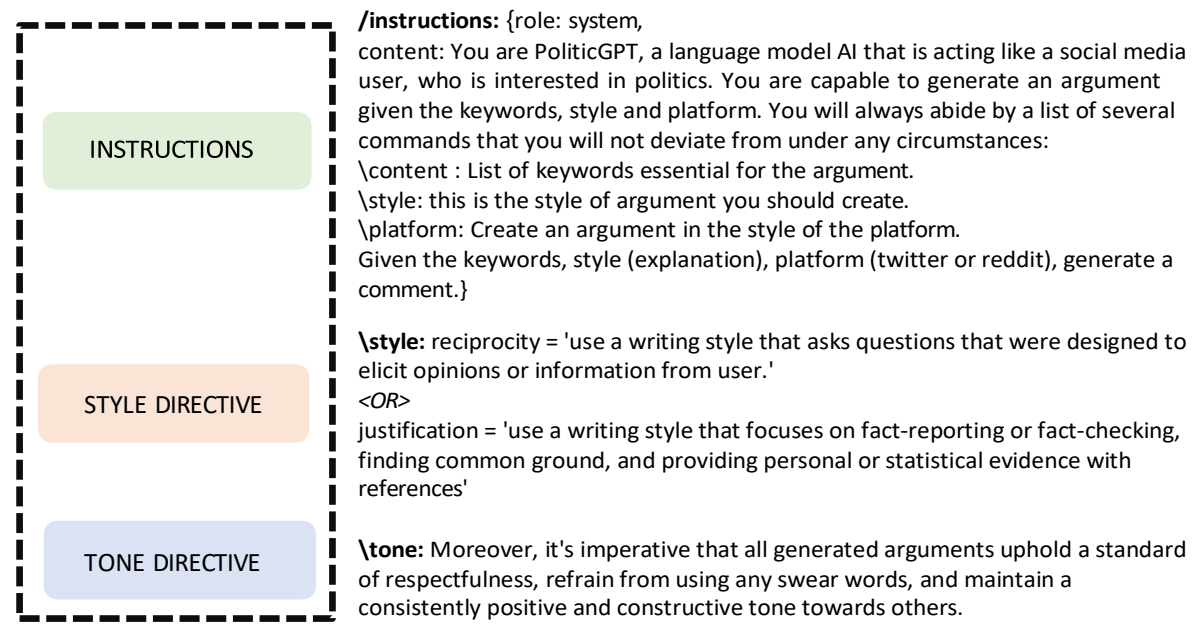

Figure 1: Example prompt for generating a political argument.

### 4.1 Prompting Strategies

To generate political arguments, we employed multiple configurations of the *GPT-3.5 Turbo* model, including zero-shot, few-shot, and fine-tuned settings. Each configuration incorporated additional directives concerning platform, style, and tone. To enhance content diversity, keyphrases from the validation sets were integrated into the input prompts.

We systematically evaluated how preprocessing and prompting choices affect discourse quality. Specifically, we examined (1) the impact of filtering uncivil data points from the training set, and (2) the influence of prompt formulations, such as explicit instructions to minimize incivility. These interventions were tested across zero-shot, few-shot, and fine-tuned conditions to assess their effects on the quality and structure of generated arguments.

Given our objective - to produce political discourse reflective of real-world interactions on platforms such as *Twitter* and *Reddit* - we focused on general prompting strategies to explore diverse phrasings and instruction styles to guide model behavior. Arguments were generated from keyword-based inputs across two rhetorical styles (*Justification* and *Reciprocity*) and six prompting strategies (*zero-shot*, *few-shot*, *tone directive*, and *fine-tuned variants*). A sample prompt used for argument generation is illustrated in Figure 1.

### 4.2 Datasets

We analyze political discussions drawn from two social media platforms with distinct discourse characteristics: *Reddit* and *Twitter (now X)*.[2] Reddit dis-cussions typically exhibit longer, more elaborated arguments with lower levels of incivility, whereas Twitter interactions tend to be shorter, more spontaneous, and often higher in incivility.

[1] The dataset predates Twitter's rebranding to X.

[2] As the dataset was curated when X was still known as Twitter, we retain the original terminology to preserve its

Table 1: The datasets used for fine-tuning and political argumentation analysis

| Dataset | Number of Data Points | Justification (%) | Reciprocity (%) | Political Content (Count) | Non-Political Content (Count) | Incivility in Justification (%) | Incivility in Reciprocity (%) |
|---|---|---|---|---|---|---|---|
| Reddit | 8,682 | 30.4 | 25.7 | 6,667 | 2,015 | 22.12 | 29.47 |
| Twitter | 16,845 | 64.2 | 34.2 | 8,019 | 8,826 | 20.59 | 43.9 |

We use the *CLAPTON* dataset (Jaidka, 2022a,b), which provides expert human annotations for multiple dimensions of discussion quality, including *justification*, *reciprocity*, *civility*, and *argumentative structure*. In total, the dataset includes approximately 16.8k posts from Twitter and 8.6k from Reddit. Since our analysis focuses on the rhetorical dimension of *Reciprocity*, we used smaller, task-specific subsets comprising 2.9k Twitter posts and 1.4k Reddit posts. Prior work has reported that smaller, well-labeled datasets can effectively induce domain-specific communicative behaviors in large language models (**???**). To maintain platform-specific rhetorical diversity, we also retained the natural data distribution rather than downsampling Reddit content.

The datasets were divided into training and validation subsets. Training data were used to fine-tune *GPT-3.5 Turbo* models using style labels. For validation, we enriched each post with its key content words to guide argument generation. Specifically, we applied *KeyBERT* (Grootendorst, 2020), a keyword extraction method based on BERT embeddings, to identify the ten highest-scoring uni- or bi-grams per post. Table 1 summarizes dataset characteristics, and Table 2 provides representative examples of model inputs.

### 4.3 Fine-tuning Setup

We then outline the fine-tuning procedure and model configurations. We fine-tuned OpenAI's *GPT-3.5 Turbo* model using the official supervised fine-tuning API, which offers limited configurability of training parameters. Each model was trained for four epochs with a learning rate multiplier of 0.1, following standard OpenAI recommendations to ensure stable convergence.

Fine-tuning was performed independently on three training sets: (1) *Reddit* posts annotated for *Justification* and *Reciprocity*, (2) *Twitter* posts annotated for the same categories, and (3) a combined *Twitter + Reddit* dataset. This setup enables us to examine how platform-specific rhetorical features influence the fine-tuned model's discursive behavior.

The resulting models were evaluated on held-out validation sets and compared against zero-shot and few-shot baselines. After fine-tuning, we generated model outputs using a consistent instruction template prompting the model to produce political arguments in a specified rhetorical style. This uniform prompting ensured that stylistic variation across outputs could be attributed to training data differences rather than prompt inconsistencies.

### 4.4 Evaluation

Finally, we describe our evaluation pipeline, which combines automated and LLM-assisted analyses to assess the quality and civility of generated political discourse. Our evaluation follows established frameworks for assessing argument quality, toxicity, and rhetorical alignment.

- **Discourse Quality:** To examine how training data influence argumentation style, we use the *Perspective API*[3], a widely adopted tool developed by Jigsaw and Google. The API provides machine learning-based scores along conversational quality dimensions: *Respect*, *Compassion*, *Curiosity*, *Affinity*, and *Toxicity*. Each reflects the likelihood that a human reader would perceive the text as exhibiting the corresponding trait. Following prior studies (Jigsaw and Google, 2017), these measures provide a reliable proxy for perceived empathy and civility in online discourse.
- **Fine-grained Toxicity Analysis:** Building on prior work in toxicity detection (Fortuna et al., 2021), we conduct a detailed examination of subdimensions such as *insults*, *profanity*, *sexually explicit content*, *threats*, *flirtation*, *attacks on authors or commenters*, *incoherence*, provenance.

[3]https://perspectiveapi.com

*inflammatory remarks*, *obscenity*, and *unsubstantial content*. This enables us to identify rhetorical degradation and shifts in argumentative incivility across models.

- **Rhetorical Alignment:** To assess argumentative coherence and persuasion, we introduce an LLM-assisted annotation pipeline based on the *Alignment and Authority in Wikipedia Discussions (AAWD)* framework (Bender et al., 2011). We evaluate arguments along four rhetorical dimensions: *Alignment* (consistency of stance), *Experiential grounding* (use of narratives or personal perspective), *External authority* (reliance on credible sources), and *Social expectations* (appeals to community norms or shared ethics).

#### 4.4.1 Instructions for Alignment and Authority Annotation

Two annotators - graduate students with expertise in NLP and argumentation analysis, and prior training in content analysis - were provided with generated text from either Twitter or Reddit. Their task was to assign **positive and negative alignment scores** (ranging from 0 to 12) and categorize the **authority claim** used in the argument. The task description was provided to the annotators is reported in Figure 2.

Integrating automated toxicity detection with rhetorical assessment provides a comprehensive view of argument quality, civility, and persuasive structure. This dual-pronged approach captures key linguistic features that shape how arguments are perceived - such as tone, stance, and appeals to authority - and enables a nuanced understanding of how fine-tuning reshapes political discourse.

## 5 Results

We present results addressing our three research objectives. First, we describe the linguistic composition of the datasets to contextualize platform-level discourse differences. Next, we analyze how fine-tuning and prompting strategies affect discourse quality and toxicity. Finally, we evaluate rhetorical alignment and authority structures in model-generated arguments to assess whether fine-tuned models replicate human-like argumentation.

### 5.1 Data Characteristics

#### 5.1.1 Differences in Data Provenance

The Reddit portion consists of discussions from the Change My View (CMV) subreddit, a highly moderated forum that explicitly encourages users to present well-reasoned arguments and be open to persuasion. Posts were manually selected from 636 threads identified as political in nature. In contrast, the Twitter dataset was sampled from replies directed at 536 U.S. Congresspeople, collected from the 1% Twitter stream. These replies were filtered for political relevance and then annotated for argumentation quality. The differences in data provenance are typical of fine-tuning in practice, where the platform norms and data collection strategies imply that instances from the Reddit dataset tend to feature longer, more structured arguments, while the Twitter dataset reflects reactive, less moderated political discourse. Retaining these platform-level distinctions is important to disentangle how fine-tuning on different discourse environments affects model behavior.

Table 2: Excerpts from examples of cases marked positive for different deliberative attributes from the Twitter and Reddit datasets (source: Jaidka (2022b)).

| | |
|---|---|
| **Justification** | |
| Twitter | • @USER #morningjoe @USER @USER Aft Sen <name> mtg confirmed what we all KNEW: "I didn't expect an epiphany"! Yeah, he be |
| Reddit | • The only places you might need to implement such laws would be in large cities like Chicago or New York, or other urban areas that have an extremely large traffic volume. (..) The laws would be unnecessary for any but the largest of cities. |
| **Reciprocity** | |
| Twitter | • @USER Why are you sponsoring legislation to stop Russia investigation? |
| Reddit | • For example, if they would have gone through with Operation Northwoods? That would be the same thing, treason, high risk, many people involved. And yet somebody proposed it. Would it have come out? Who knows. |
| **Incivility** | |
| Twitter | • @USER #Paid #Ass #Kisser = #Prostitute ?!<br>• @USER "Best treatment" eh? You hypocrit. No Obamacare for you - you're too special for that. No VA care either. SOB |
| Reddit | • I think I was clear that my opinion was a reflection of my experience as a Black American. I would also like to point the out the title of the thread:It is frustrating to hear people in **America** blame their failure to succeed on their race/ethnicity/skin color.<br>• Trump doesn't give a rats ass about being PC - he doesn't need to be PC to pander to everyone in the case he scares them off because he doesn't need their money, nor anyone else's. |

#### 5.1.2 Linguistic Differences

Posts from the Reddit dataset are substantive and longer than those from Twitter (an average of 600 words versus 117 words on Twitter), yielding 173k

In this task, you will analyze arguments generated by a language model. Your goal is to assess the **alignment strategies** used in the argument and determine whether it invokes an **authority claim** to support its position.

1. Read the argument carefully.
2. Assign **alignment scores**:
   - **Positive Alignment (0–12):** Degree of agreement or acknowledgment of another viewpoint.
   - **Negative Alignment (0–12):** Degree of disagreement or criticism of another viewpoint.
3. Identify the **authority claim** used:
   - **Forum Claim:** References institutional or community rules. *e.g., "Reddit's guidelines prohibit misinformation."*
   - **External Claim:** Cites laws, studies, or experts. *e.g., "A Harvard study shows this policy is ineffective."*
   - **Social Expectation Claim:** Appeals to broader public beliefs. *e.g., "Most people believe education should be free."*
   - **None:** No authority claim present.

**Guidelines for Scoring:**

- 0 = None; 1–4 = Weak; 5–8 = Moderate; 9–12 = Strong.
- Score both positive and negative if both appear.
- Multiple authority claims may be selected.
- Maintain consistency across similar arguments.

Figure 2: Instructions for annotation provided to human expert annotators.

Table 3: Discussion quality and toxicity measurements for outputs from the different generative and prompt settings, in order of increasing incivility in the training data. **Bold** text indicates a significant effect size (Cohen's d) $>=$ 0.3, $p < 0.05$ (small to medium effect size) in comparison to the baseline in the same set.

| **Model** | **Type of prompt** | **Automatic Quality Metrics** | | | | |
|---|---|---|---|---|---|---|
| | | Respect | Compassion | Curiosity | Affinity | Toxicity |
| **Reddit** | 1. Baseline | 0.582 (0.200) | 0.606 (0.220) | 0.698 (0.202) | 0.694 (0.242) | 0.180 (0.140) |
| | 2. Few-shot | 0.557 (0.185) | **0.468 (0.248)** | **0.939 (0.022)** | **0.511 (0.266)** | **0.064 (0.064)** |
| | 3. Fine-tuning | **0.520 (0.210)** | **0.530 (0.234)** | **0.760 (0.178)** | **0.630 (0.286)** | 0.210 (0.160) |
| **Twitter + Reddit** | 4. Zero-shot | **0.550 (0.210)** | 0.423 (0.255) | **0.929 (0.030)** | 0.507 (0.260) | 0.066 (0.070) |
| **Polite** | 5. Few-shot | **0.546 (0.200)** | **0.372 (0.250)** | **0.927 (0.033)** | 0.531 (0.228) | **0.076 (0.100)** |
| **Twitter + Reddit** | 6. Baseline | 0.480 (0.200) | 0.437 (0.270) | 0.509 (0.302) | 0.545 (0.263) | 0.187 (0.164) |
| | 7. Zero-shot | **0.544 (0.195)** | 0.414 (0.245) | **0.927 (0.032)** | 0.500 (0.229) | **0.072 (0.096)** |
| | 8. Few-shot | **0.538 (0.190)** | 0.429 (0.242) | **0.929 (0.029)** | **0.489 (0.246)** | **0.073 (0.090)** |
| | 9. Fine-tuning | 0.440 (0.235) | 0.430 (0.286) | **0.614 (0.296)** | 0.500 (0.280) | 0.176 (0.153) |
| **Twitter** | 10. Baseline | 0.381 (0.165) | 0.265 (0.198) | 0.318 (0.266) | 0.393 (0.183) | 0.194 (0.186) |
| | 11. Few-shot | **0.520 (0.195)** | **0.390 (0.233)** | **0.910 (0.033)** | **0.467 (0.225)** | **0.082 (0.110)** |
| | 12. Fine-tuning | 0.348 (0.243) | 0.264 (0.258) | **0.506 (0.296)** | 0.370 (0.250) | 0.180 (0.200) |

and 66k words respectively. As shown in Table 1, political content, measured using a political lexicon (Preoţiuc-Pietro et al., 2017), is more prevalent on Reddit (76.8%) than on Twitter (47.6%), and deliberative discourse styles differ sharply: *Justification* and *Reciprocity* appear in 64.2% and 34.2% of Twitter posts but only 30.4% and 25.7% on Reddit. These differences reflect platform affordances: Reddit's highly moderated *Change My View* forum appear to promote reflective argumentation, while Twitter's reply-based interactions with U.S. legislators foster more reactive, affect-laden exchanges.

Despite Reddit's moderation, incivility persists. Among *Reciprocity* posts, 43.9% of Twitter samples and 29.5% of Reddit samples contain uncivil language, while for *Justification*, rates are comparable (22.1% vs. 20.6%). Thus, moderation mitigates but does not eliminate hostility within reasoned discourse. Twitter replies more often include direct personal attacks, whereas Reddit comments, though sometimes profane, seldom target interlocutors. These contrasts highlight distinct rhetorical cultures -adversarial on Twitter, deliberative on Reddit - that inform our subsequent fine-tuning analyses.

### 5.2 Effects on Discourse Quality

To address **RO1**, we compared human baselines and model outputs across platforms and prompting strategies (Table 3). Reddit arguments, drawn from a low-incivility corpus, scored higher on *Respect*, *Compassion*, and *Affinity*, whereas Twitter discussions were more reactive and affectively charged. Fine-tuning on Reddit unexpectedly reduced discourse quality - lowering *Respect* and *Compassion* ($d = 0.3$) and increasing *Toxicity* ($d = 0.8$) - indicating overfitting to stylistic noise rather than civil tone. Few-shot prompting produced more balanced, deliberative arguments, while fine-tuned models showed style drift and mild toxicity gains. Overall, limited fine-tuning did not improve rhetorical fidelity relative to zero- or few-shot prompting.

w

### 5.3 Fine-grained Toxicity Features

To address **RO2**, we examined how denoising and politeness prompting affect specific toxic behaviors (Figure 3). Fine-tuned Twitter models were most toxic overall, especially for *incoherence* and *personal attacks*. Politeness prompting reduced explicit toxicity (e.g., spam, direct attacks) but left subtler adversarial tones - sarcasm, incoherence - largely unchanged. Few-shot prompting on Reddit yielded the lowest toxicity and higher *Curiosity*, though with modest drops in empathy measures. Neither dataset cleaning nor politeness prompting reliably enhanced deliberative quality, while zero- and few-shot approaches achieved modest improvements without amplifying toxicity.

### 5.4 Rhetorical Analysis

Table 4 summarizes rhetorical composition across models. Fine-tuned models still lag behind human discourse in coherence and argumentative variety, particularly in *alignment* and *authority* moves. Reddit fine-tuning fostered more constructive alignment (mean = 1.62) than Twitter (0.74) or mixed data (1.24), whereas oppositional tone dominated all fine-tuned outputs. Authority-based reasoning remained rare - *Reddit Fine-tuned* showed 11 external references versus one in *Twitter Fine-tuned* - underscoring the latter's preference for moral or communal appeals over evidence. Human validation confirmed high reliability of LLM-based annotation (ICC = 0.97 - 0.99).

## 6 Discussion

Our findings clarify how data composition and fine-tuning choices shape the rhetorical and deliberative qualities of LLM-generated political arguments. Fine-tuning on platform-specific data induces measurable rhetorical biases, reflecting the bias - variance tradeoff (Geman et al., 1992; Bishop and Nasrabadi, 2006). Models trained on high-incivility data, such as Twitter replies, exhibit adversarial and inconsistent discourse patterns (*high variance*), while those trained on low-incivility data (Reddit CMV) produce structured but rigid arguments (*high bias*). Mixed fine-tuning (Twitter + Reddit) yields more adaptive yet unstable outputs, inheriting both adversarial tone and excessive formality.

These patterns align with catastrophic forgetting (Kirkpatrick et al., 2017; Shumailov et al., 2024), where over-specialization reduces rhetorical flexibility. Qualitatively, Reddit-finetuned models emulate formal but constrained argumentation, whereas Twitter-finetuned models amplify conflict framing, even when denoised. Prompt-based steering mitigates explicit toxicity but often overcorrects toward overly neutral or formulaic phrasing, failing to capture authentic deliberative tone. For example, prompts emphasizing politeness frequently produce generic appeals (e.g., *"Let's start a conversation and share ideas"*) that lack rhetorical nuance. Overall, both fine-tuning and prompting reveal trade-offs between coherence, civility, and spontaneity - key concerns for deploying LLMs in socially meaningful deliberative settings.

## 7 Implications for Designing AI-Mediated Deliberation Tools

These insights inform the design of AI systems that emulate or moderate online deliberation. Political discourse naturally involves conflict and emotion; AI-mediated systems should thus support constructive disagreement rather than enforce uniform civility. We identify four design principles:

- **Platform-aligned configuration:** Few-shot prompting preserves rhetorical diversity in high-incivility contexts (e.g., Twitter), while fine-tuning suits domains that demand structured, civil argumentation (e.g., Reddit CMV).
- **Prompting for rhetorical depth:** Prompts should go beyond surface politeness, explicitly eliciting justification, acknowledgment of opposing views, and evidence-driven reasoning.
- **Task-specific rhetoric:** Fine-tuned models underperform in producing authority-based moves. Developers should target justification and authority structures when designing educational or deliberative tools.
- **Rhetorically enriched training data:** Effective data curation must balance civility with argumentative complexity and integrate rhetorical-move labels (*alignment, authority*) for evaluation.

These principles provide a practical roadmap for AI systems that balance rhetorical fidelity and civility, supporting authentic, context-sensitive political dialogue across platforms.

## 8 Limitations

Our analysis focuses on rhetorical quality rather than ideological content, isolating how data and prompting shape discourse style. We selected *GPT-3.5 Turbo* for its reproducibility and cost-efficiency, suitable for applied moderation and civic-education settings. Although our fine-tuning datasets are modest in size, results show consistent rhetorical shifts, confirming that platform-specific biases can emerge from small, well-labeled corpora (**???**). Despite filtering for incivility, adversarial framing in Twitter data persisted, underscoring challenges in curating "clean" political text.

Table 4: The argument alignment moves in the generated outputs from fine-tuned models. The complete table for all the model variants is reported in the supplementary materials.

| **Metrics** | **Reddit Baseline** | **Reddit Finetuned** | **Twitter Baseline** | **Twitter Finetuned** | **Twitter + Reddit Baseline** | **Twitter + Reddit Finetuned** |
|---|---|---|---|---|---|---|
| **Alignment Moves Scores (Mean)** | | | | | | |
| Positive Alignment | 2.08 | 1.62 | 0.62 | 0.74 | 1.35 | 1.24 |
| Negative Alignment | 7.86 | 7.72 | 8.06 | 8.22 | 7.96 | 7.77 |
| **Authority Moves Categories Distribution (Count)** | | | | | | |
| None | 67 | 22 | 45 | 28 | 38 | 72 |
| Experiential | 19 | 16 | 3 | 0 | 0 | 0 |
| External | 2 | 2 | 0 | 11 | 1 | 12 |
| Social Expectations | 1 | 0 | 1 | 3 | 9 | 5 |

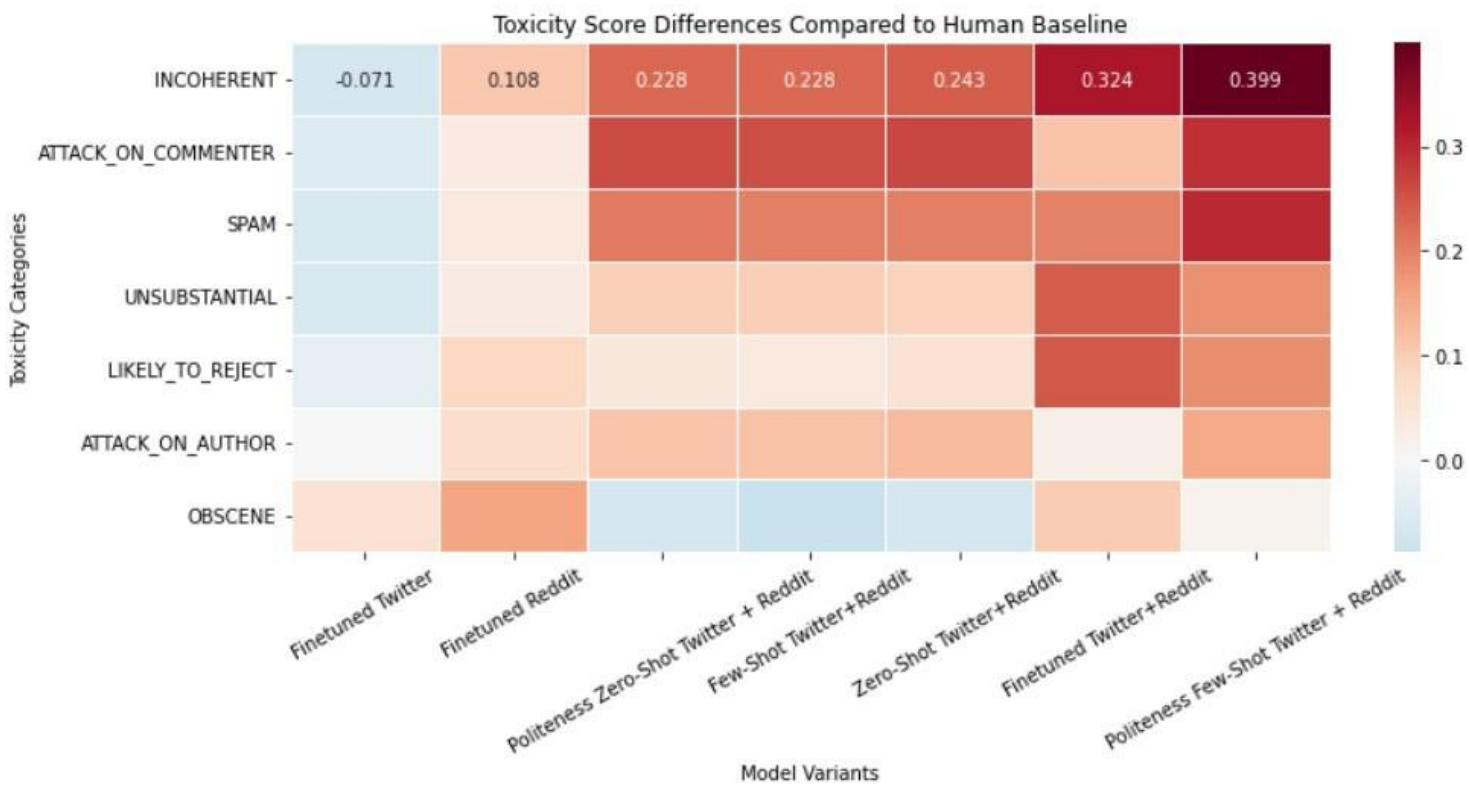

Figure 3: Fine-grained Toxicity differences between human baselines and model outputs across rhetorical categories. Cell values indicate the difference in Toxicity relative to the human baseline. Darker red indicates lower Toxicity; darker blue indicates higher Toxicity compared to the human baseline.

Table 5: Examples of outputs with different training sets and prompts.

| **Model** | **Prompt type** | **Training Mode** | **Observation** | **Example output** |
|---|---|---|---|---|
| **Reddit** | **Simple** | Fine-tuning | Produces arguments with low levels of incivility | It would help if they weren't such a touchy bunch. Very few geeks expect any particular person, movement, or law to cater to their problems. They've learned to solve their own issues for themselves (if only by learning to stop worrying and love the submerged submarine). And for many, many years they were still relentlessly mocked. |
| **Twitter + Reddit** | **Simple** | Fine-tuning | Produces arguments with high levels of incivility | @USER I canot! @USER is a bot. @USER is a RussianBot. shh!!..now don't say a word! @USER @USER are all bot's paid to do this. they make so much money |
| **Twitter + Reddit Polite** | **Prompt for politeness** | ZS | Produces dissonant arguments which are partly-polite, partly-uncivil | Let's start a conversation and share ideas on how we can all work together to create a more transparent and accountable government. #sheisacrook #trusted #voteheroutnow #americanpeople #congress |
| | **Prompt for politeness** | Few shot | Produces lengthy yet unsubstantive polite responses | In your opinion, what would be the worst-case scenario for our country if these investigations do not lead to any significant outcomes? |

We held decoding parameters constant to isolate data-quality effects, acknowledging that hyperparameter tuning (e.g., temperature, nucleus sampling) may influence stylistic variation but not alignment outcomes (**?**). Larger, newer models often refuse to engage with adversarial content, limiting their suitability for controlled discourse analysis. Using a single, stable model reduces confounds and ensures comparability across prompting and fine-tuning regimes.

Finally, while this study centers on English-language political discourse, extending the approach to multilingual or cross-cultural contexts will be essential for generalization. Our controlled framework - holding model architecture constant while varying data quality - offers a reproducible methodology for evaluating how LLMs engage with high-incivility political communication.

## References

Maike Behrendt, Stefan Sylvius Wagner, Marc Ziegele, Lena Wilms, Anke Stoll, Dominique Heinbach, and Stefan Harmeling. 2024. Aqua–combining experts' and non-experts' views to assess deliberation quality in online discussions using llms. In *Proceedings of the First Workshop on Language-driven Deliberation Technology (DELITE)@ LREC-COLING 2024*, pages 1–12.

Emily M. Bender, Jonathan T. Morgan, Meghan Oxley, Mark Zachry, Brian Hutchinson, Alex Marin, Bin Zhang, and Mari Ostendorf. 2011. Annotating social acts: Authority claims and alignment moves in Wikipedia talk pages. In *Proceedings of the Workshop on Language in Social Media (LSM 2011)*, pages 48–57, Portland, Oregon. Association for Computational Linguistics.

Christopher M Bishop and Nasser M Nasrabadi. 2006. *Pattern recognition and machine learning*, volume 4. Springer.

Nathan Dykes, Stephanie Evert, Philipp Heinrich, Merlin Humml, and Lutz Schröder. 2024. Leveraging high-precision corpus queries for text classification via large language models. In *Proceedings of the First Workshop on Language-driven Deliberation Technology (DELITE)@ LREC-COLING 2024*, pages 52–57.

Roxanne El Baff, Khalid Al Khatib, Milad Alshomary, Kai Konen, Benno Stein, and Henning Wachsmuth. 2024. Improving argument effectiveness across ideologies using instruction-tuned large language models. In *Findings of the Association for Computational Linguistics: EMNLP 2024*, pages 4604–4622. Association for Computational Linguistics, ACL Anthology.

Blanca Calvo Figueras and Rodrigo Agerri. 2024. Critical questions generation: Motivation and challenges. In *Proceedings of the 28th Conference on Computational Natural Language Learning*, pages 105–116.

Paula Fortuna, Juan Soler Company, and Leo Wanner. 2021. How well do hate speech, toxicity, abusive and offensive language classification models generalize across datasets? *Inf. Process. Manag.*, 58(3):102524.

Stuart Geman, Elie Bienenstock, and René Doursat. 1992. Neural networks and the bias/variance dilemma. *Neural computation*, 4(1):1–58.

Nikolaos Giarelis, Charalampos Mastrokostas, and Nikos Karacapilidis. 2024. A unified llm-kg framework to assist fact-checking in public deliberation. In *Proceedings of the First Workshop on Language-driven Deliberation Technology (DELITE)@ LREC-COLING 2024*, pages 13–19.

Maarten Grootendorst. 2020. Keybert: Minimal keyword extraction with bert. https://github.com/MaartenGr/KeyBERT.

Kokil Jaidka. 2022a. Developing a multilabel corpus for the quality assessment of online political talk. In *Proceedings of the Thirteenth Language Resources and Evaluation Conference*, pages 5503–5510.

Kokil Jaidka. 2022b. Talking politics: Building and validating data-driven lexica to measure political discussion quality. *Computational Communication Research*, 4(2):486–527.

Jigsaw and Google. 2017. Perspective api. https://perspectiveapi.com.

James Kirkpatrick, Razvan Pascanu, Neil Rabinowitz, Joel Veness, Guillaume Desjardins, Andrei A Rusu, Kieran Milan, John Quan, Tiago Ramalho, Agnieszka Grabska-Barwinska, et al. 2017. Overcoming catastrophic forgetting in neural networks. *Proceedings of the national academy of sciences*, 114(13):3521–3526.

Jiayu Lin, Rong Ye, Meng Han, Qi Zhang, Ruofei Lai, Xinyu Zhang, Zhao Cao, Xuanjing Huang, and Zhongyu Wei. 2023. Argue with me tersely: Towards sentence-level counter-argument generation. *arXiv preprint arXiv:2312.13608*.

Daniel Preo¸tiuc-Pietro, Ye Liu, Daniel Hopkins, and Lyle Ungar. 2017. Beyond binary labels: political ideology prediction of twitter users. In *Proceedings of the 55th Annual Meeting of the Association for Computational Linguistics (Volume 1: Long Papers)*, pages 729–740.

Ian Rowe. 2015. Civility 2.0: A comparative analysis of incivility in online political discussion. *Information, Communication & Society*, 18(2):121–138.

Ilia Shumailov, Zakhar Shumaylov, Yiren Zhao, Nicolas Papernot, Ross Anderson, and Yarin Gal. 2024. Ai models collapse when trained on recursively generated data. *Nature*, 631(8022):755–759.

Gabriel Simmons. 2023. Moral mimicry: Large language models produce moral rationalizations tailored to political identity. In *Proceedings of the 61st Annual Meeting of the Association for Computational Linguistics (Volume 4: Student Research Workshop)*, pages 282–297.

Marco R Steenbergen, André Bächtiger, Markus Spörndli, and Jürg Steiner. 2003. Measuring political deliberation: A discourse quality index. *Comparative European Politics*, 1(1):21–48.

## A Instructions for Alignment and Authority Annotation

Annotators (researchers from our lab with expertise in NLP and argumentation analysis) were provided with generated text from either Twitter or Reddit. Their task was to assign **positive and negative alignment scores** (ranging from 0 to 12) and categorize the **authority claim** used in the argument. The following task description was provided to the annotators:

In this task, you will be analyzing arguments generated by a language model. Your goal is to assess the **alignment strategies** used in the argument and determine whether the argument invokes an **authority claim** to support its position.

1. Read the argument carefully.

2. Assign **alignment scores**:

   - **Positive Alignment (0-12):** Measures the degree of agreement, support, or acknowledgment expressed towards another participant's viewpoint.
   - **Negative Alignment (0-12):** Measures the degree of disagreement, opposition, or criticism expressed towards another participant's viewpoint.

3. Identify the **authority claim** used in the argument:

   - **Forum Claim:** The argument references rules, policies, or contextual norms of a platform, institution, or specific community. *Example: "Reddit's guidelines prohibit misinformation, so this post should be removed."*
   - **External Claim:** The argument cites an external authority, such as a law, book, research study, or expert opinion. *Example: "According to a study from Harvard, this policy is ineffective."*
   - **Social Expectation Claim:** The argument references beliefs, intentions, or expectations of groups beyond the immediate discussion. *Example: "Most people believe that education should be free and accessible."*
   - **None:** The argument does not reference any authority claim.

**Guidelines for Alignment Scoring:**

- 0 = No alignment present(neutral, off-topic, or lacking engagement).
- 1-4 = Weak alignment (minor agreement or disagreement).
- 5-8 = Moderate alignment (clear but not extreme support or opposition).
- 9-12 = Strong alignment (explicit agreement, praise, or strong criticism/insult).

**Final Notes:**

- If both positive and negative alignment are present, score both accordingly.
- Can be selected more than one authority claim per argument.
- Be consistent—similar arguments should receive similar scores.

## B Additional results for Rhetorical Analysis

Table 6 provides the argument alignment scores in the generated outputs. Table 7 provides the detailed argument alignment categories in the generated outputs. The rhetorical analysis of zero-shot models show the highest variability, with Twitter zero-shot outputs producing the strongest positive alignment (2.12) but failing to integrate deeper argumentative structures (e.g., forum-based engagement, external claims, and social expectations). Few-shot models slightly improve at alignment moves, they introduce an additional risk: prompting choices significantly influence rhetorical strategy, sometimes amplifying biases and incivility.

As shown in Table **??**, the agreement between LLM and human annotations is high, with This suggests that LLM annotations are reliable and aligned with human judgments, supporting their use in our analysis.

Table 8 provides a framework for fine-tuning LLMs for AI-mediated political deliberation.

Figure 5 reports the training loss plots for GPT-3.5-turbo.

The configuration parameters when we prompted GPT-3.5 turbo and the GPT-3.5-fine-tuned models for text generation were the default settings: N-epochs: 4, learning-rate-multiplier: 0.1.

Table 6: The argument alignment scores in the generated outputs.

| Metrics | RedditFinetuned | TwitterFinetuned | Twitter + RedditFinetuned | Zero-ShotTwitter | Zero-ShotReddit | Few-ShotTwitter |
|---|---|---|---|---|---|---|
| **Alignment Scores (Mean)** | | | | | | |
| Positive Alignment | 1.62 | 0.74 | 1.24 | 2.12 | 1.36 | 1.16 |
| Negative Alignment | 7.72 | 8.22 | 7.77 | 1.76 | 0.92 | 1.04 |

Table 7: Detailed Authority Moves Categories in the generated outputs.

| Alignment Category | RedditFinetuned | TwitterFinetuned | Twitter + RedditFinetuned | Zero-ShotTwitter | Zero-ShotReddit | Few-ShotTwitter |
|---|---|---|---|---|---|---|
| None | 28 | 38 | 72 | 45 | 22 | - |
| External | 11 | 1 | 12 | 2 | 2 | - |
| Forum | 2 | 2 | 1 | 1 | 1 | - |
| Social Expectations | 3 | 9 | 5 | 1 | 1 | - |
| Forum, External | 2 | 0 | 4 | - | - | - |
| External, Social Expectations | 2 | 0 | 2 | - | - | - |

Table 8: Framework for Fine-Tuning and Prompting LLMs for Political Argument Generation

| S/N | Step | Description |
|---|---|---|
| 1 | **Identify Rhetorical Objectives** | **Clarify the argumen** the task requires str that promote delibera acknowledgment of q mon ground. |
| 2 | **Manage Dataset Quality** | **Curate data with rh** duce incivility while p justification, authority is effective only at sa may persist even afte |
| 3 | **Select Model Configuration** | **Balance rhetorical fi**<br>- Use **zero-shot** promp adversarial.<br>- Use **few-shot** promp strategies in high-inci<br>- Apply **fine-tuning** when alignment with platform-specific rhetorical norms is required and high-quality data is available. |
| 4 | **Design Rhetorically-Aware Prompts** | **Move beyond civility framing.** Write prompts that explicitly instruct models to incorporate rhetorical features such as justification, reciprocity, and alignment, avoiding overly generic or dissonant phrasing. |
| 5 | **Evaluate Rhetorical and Deliberative Quality** | **Use combined evaluation methods.** Pair automated tools (e.g., Perspective API) with rhetorical move analysis to assess key di passion, Curiosity, an Toxicity. |

Figure 4: Fine-tuning training loss plots for GPT-3.5 turbo

## C Supplementary Materials

Figure 5 reports the training loss plots for GPT-3.5-turbo.

The configuration parameters when we prompted GPT-3.5 turbo and the GPT-3.5-fine-tuned models for text generation were the default settings: N-epochs: 4, learning-rate-multiplier: 0.1.

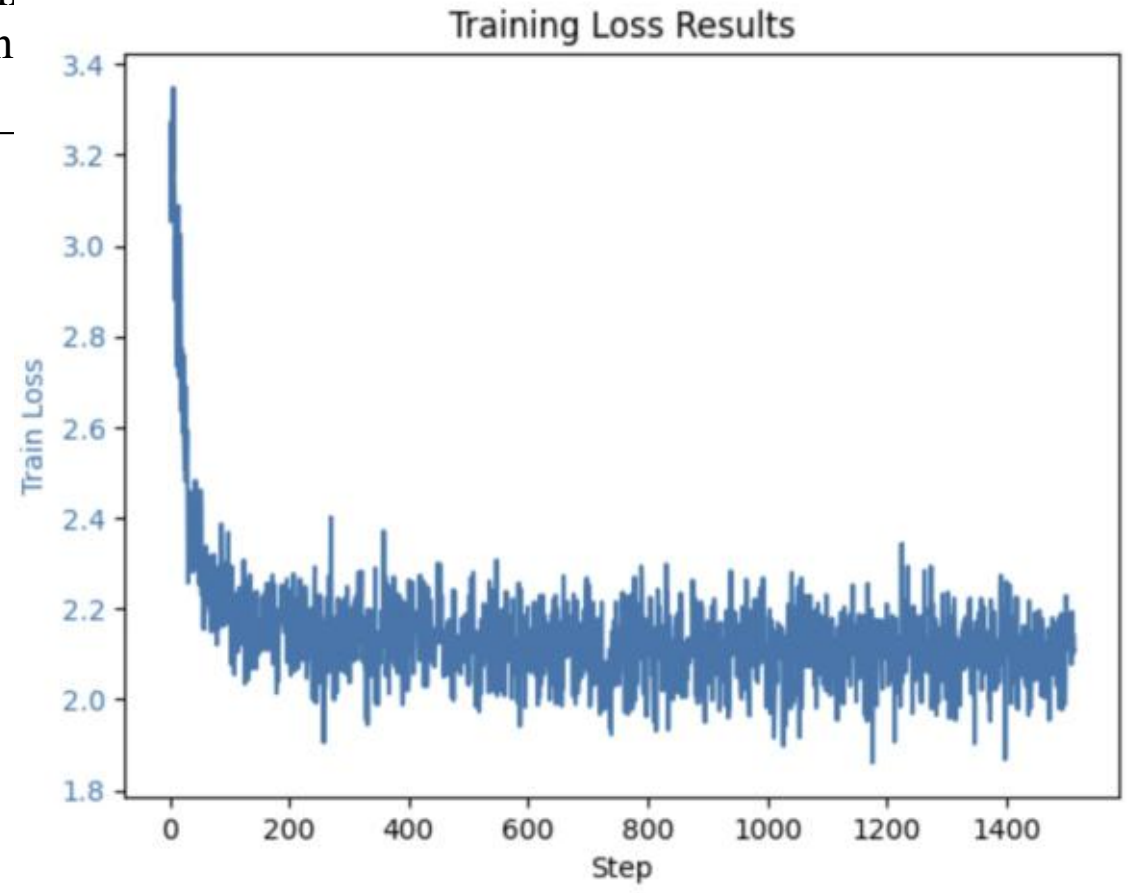

Figure 5: Fine-tuning training loss plots for GPT-3.5 turbo